\documentclass{article}
\usepackage{stywhispers,amsmath,epsfig}
\usepackage{hyperref}
\usepackage{cleveref}
\usepackage{siunitx}
\usepackage{booktabs}
\usepackage{subcaption}

\title{HyDe: The First Open-Source, Python-based, GPU-accelerated Hyperspectral Denoising Package}
\name{
    Daniel Coquelin\textsuperscript{1,2},
    Behnood Rasti\textsuperscript{3},
    Markus Götz\textsuperscript{1,2},
    Pedram Ghamisi\textsuperscript{3},
    Richard Gloaguen\textsuperscript{3},
    and Achim Streit\textsuperscript{2}
\thanks{Thanks to XYZ agency for funding.}
}
\address{
\textsuperscript{1} Helmholtz AI\\
\textsuperscript{2} Karlsruhe Institute of Technology (KIT) \\  Hermann-von-Helmholtz-Platz 1 76344, Eggenstein-Leopoldshafen, Germany \\
\textsuperscript{3} Helmholtz-Zentrum Dresden-Rossendorf (HZDR)\\ Helmholtz Institute Freiberg for Resource Technology, Chemnitzer Str. 40, D-09599 Freiberg, Germany
}
\begin{document}
%
\maketitle
\begin{abstract}
As with any physical instrument, hyperspectral cameras induce different kinds of noise in the acquired data. 
Therefore, Hyperspectral denoising is a crucial step for analyzing hyperspectral images (HSIs). 
Conventional computational methods rarely use GPUs to improve efficiency and are not fully open-source. 
Alternatively, deep learning-based methods are often open-source and use GPUs, but their training and utilization for real-world applications remain non-trivial for many researchers.
Consequently, we propose HyDe: the first open-source, GPU-accelerated Python-based, hyperspectral image denoising toolbox, which aims to provide a large set of methods with an easy-to-use environment.
HyDe includes a variety of methods ranging from low-rank wavelet-based methods to deep neural network (DNN) models. 
HyDe's interface dramatically improves the interoperability of these methods and the performance of the underlying functions.
In fact, these methods maintain similar HSI denoising performance to their original implementations while consuming nearly ten times less energy.
Furthermore, we present a method for training DNNs for denoising HSIs which are not spatially related to the training dataset, i.e., training on ground-level HSIs for denoising HSIs with other perspectives including airborne, drone-borne, and space-borne.
To utilize the trained DNNs, we show a sliding window method to effectively denoise HSIs which would otherwise require more than 40 GB.
The package can be found at: \url{https://github.com/Helmholtz-AI-Energy/HyDe}.
\end{abstract}
\section{Introduction}
\label{sec:intro}

Given the current climate and energy crises, energy efficient computational methods are crucial.
However, the benefits of efficient methods are lost if they cannot be easily used.
To this end, we introduce HyDe, the first open-source, Python-based, GPU-accelerated, hyperspectral denoising package\footnote{https://github.com/Helmholtz-AI-Energy/HyDe}.
The purpose of HyDe is to make HSI denoising and analysis as energy efficient as possible with a user-friendly interface. 

Over time, the equipment used to capture hyperspectral images (HSIs) degrades, increasing the amount of noise in the acquired data.
Therefore, denoising is a critical step when working with HSIs.
The most basic type of noise is additive noise, e.g. Gaussian blurring. 
There are many other types of noise.
For example, the deadline noise refers to dead columns of pixels and the impulse noise effects random pixels that are stuck fully on or off.

Denoising methods include full- and low-rank conventional methods, and deep learning approaches.
These methods are often designed to take advantage of the unique properties of HSIs that are not present in standard three-channel images.
Although these methods are effective, most of the conventional methods only make use of CPUs.
GPUs have shown to be faster and more energy efficient than CPUs with minimal accuracy loss~\cite{qasaimeh2019comparing}. 
Contrastively, deep neural network (DNN) methods have shown great successes recently but often require expensive tuning~\cite{evooptim}.

Currently, HyDe includes 12 HSI denoising methods and the utility functions necessary for their use.
These methods range from conventional full- and low-rank methods to DNNs and are implemented to promote ease-of-use.
HyDe offers pre-trained models of all DNN methods.

In this work, we show that HyDe is often faster and more energy efficient than the reference implementations without great sacrifices to accuracy, while also collecting all approaches into a singular toolbox with a unified interface.
We will give an overview of the included high-level functions. 
Then, we will show experiments on the efficacy of HyDe's implementations. 

\section{HyDe - Methods}
\label{sec:methods}

\begin{figure*}[ht]
  \centering
  \begin{subfigure}[b]{0.24\textwidth}
    \centering
    \includegraphics[width=\textwidth]{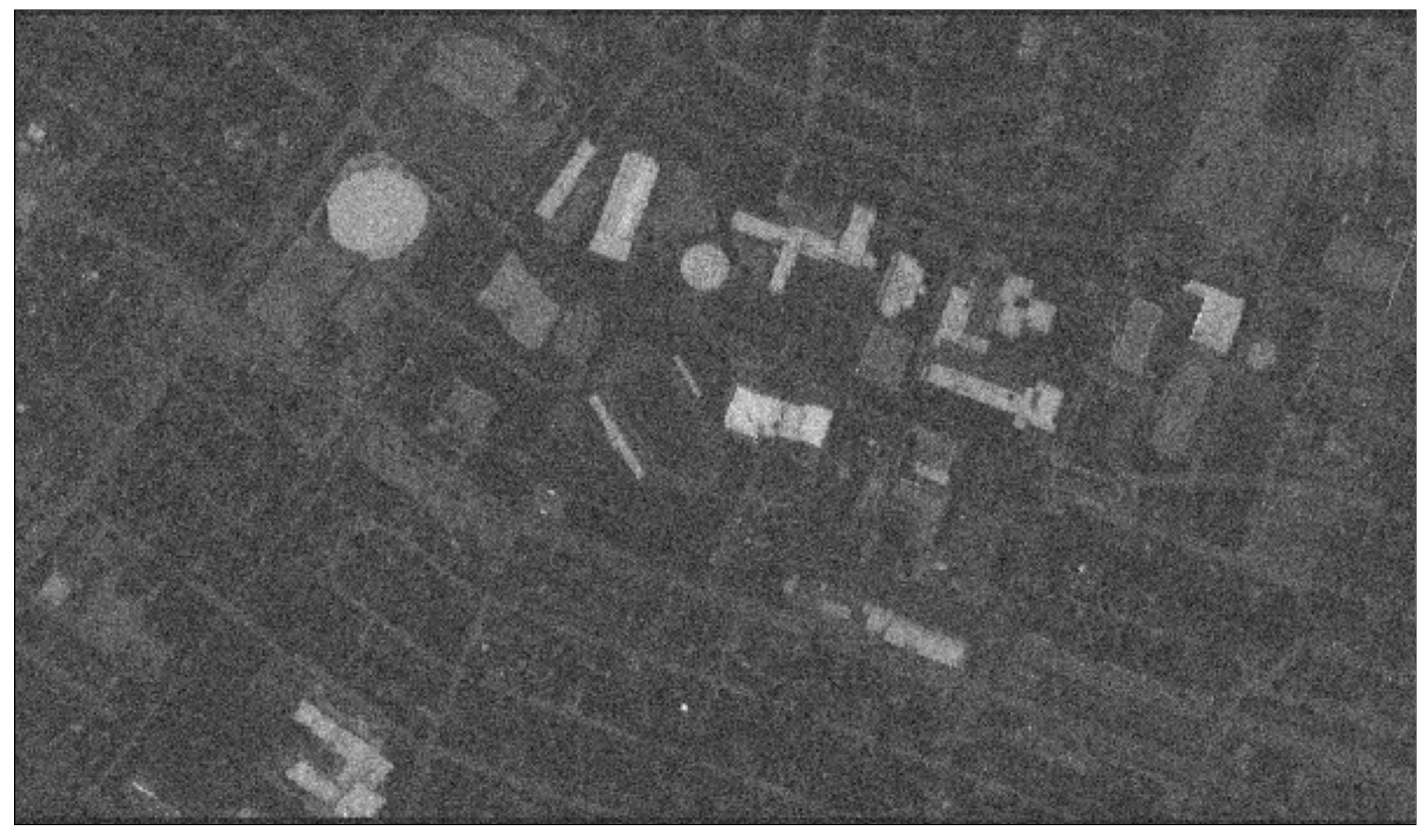}
    \caption{Noisy HSI}
    \label{fig:noisy}
  \end{subfigure}
  \hfill
  \begin{subfigure}[b]{0.24\textwidth}
    \centering
    \includegraphics[width=\textwidth]{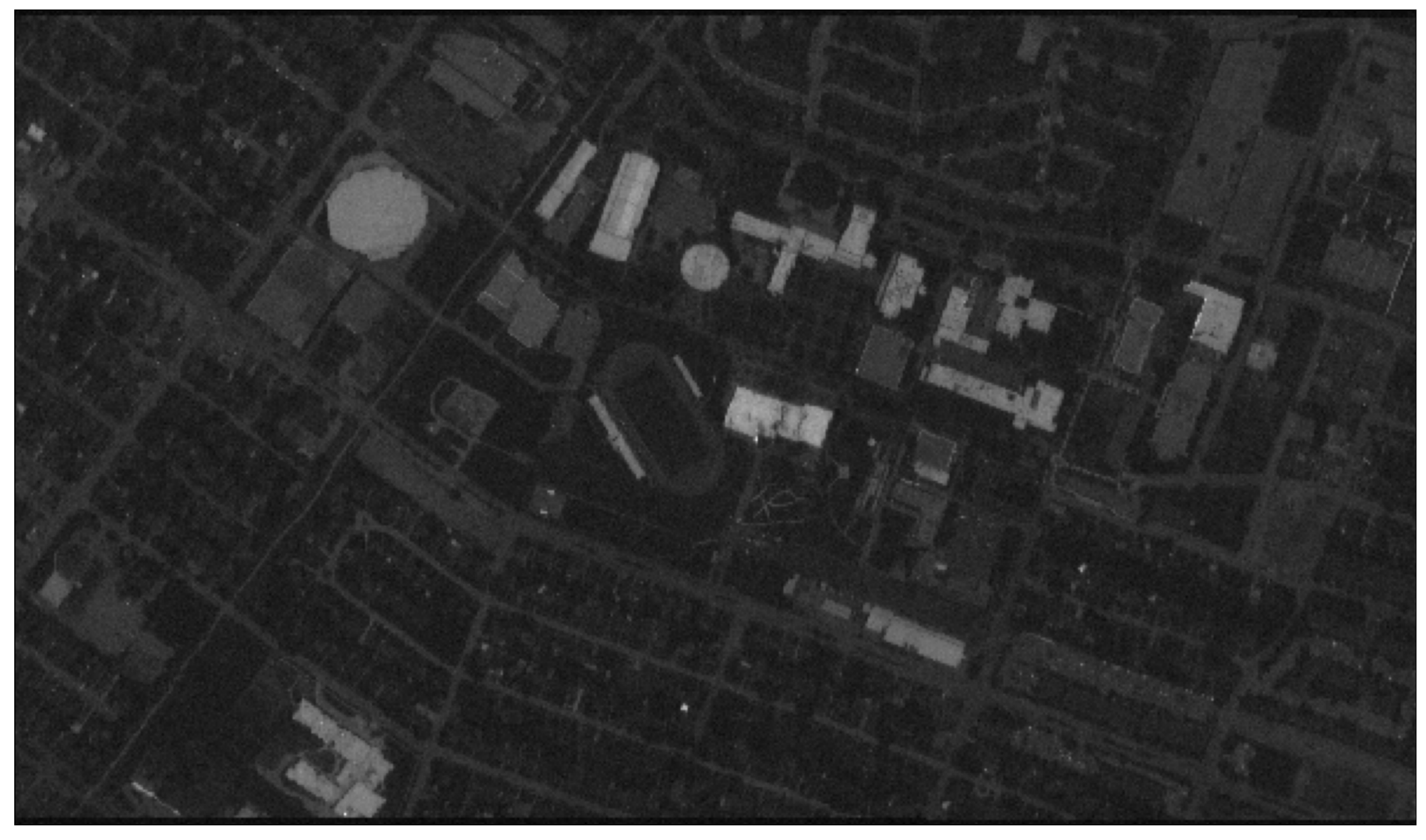}
    \caption{HyDe HyRes}
    \label{fig:hyde-hyres}
  \end{subfigure}
  \hfill
  \begin{subfigure}[b]{0.24\textwidth}
   \centering
   \includegraphics[width=\textwidth]{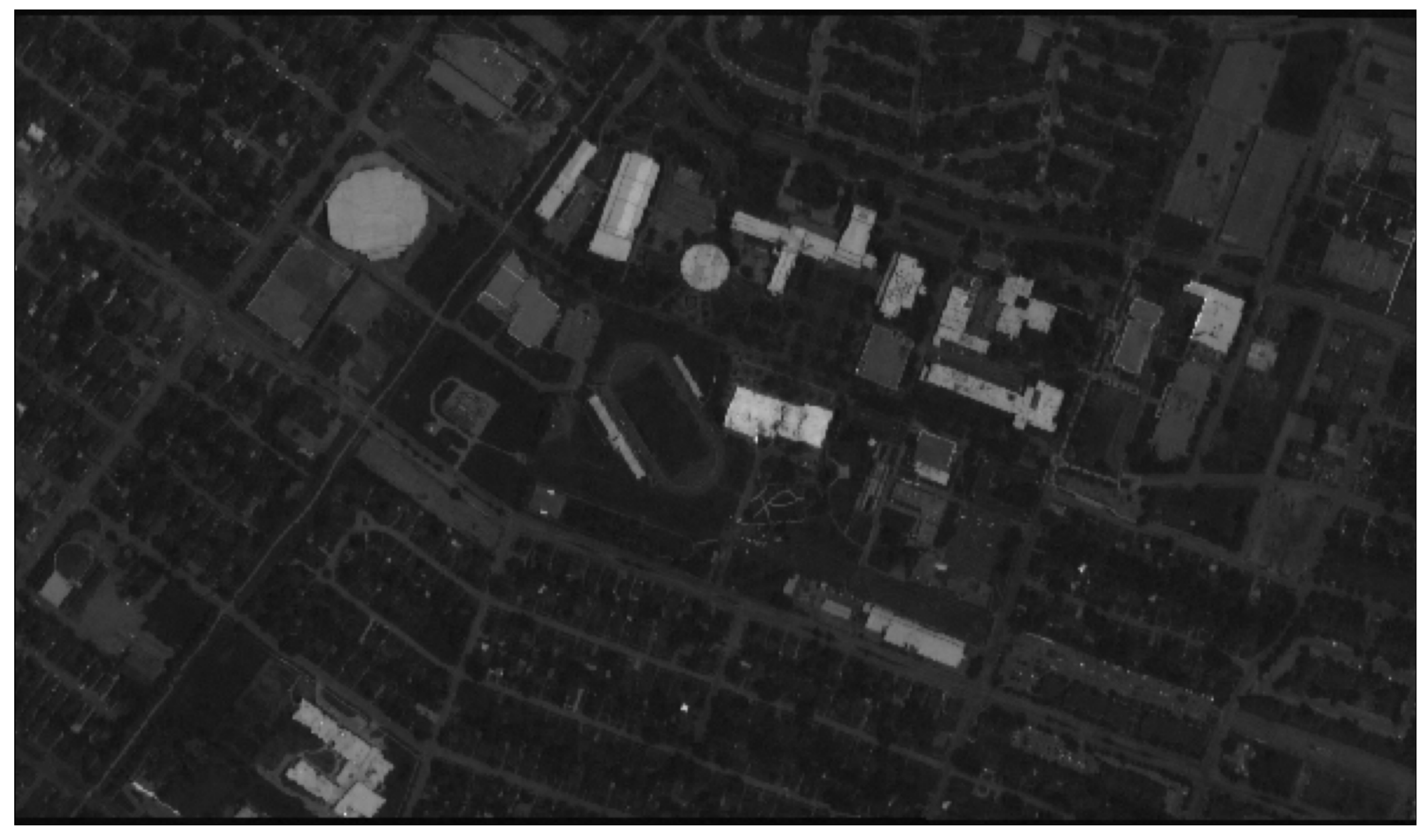}
   \caption{Original HyRes}
   \label{fig:matlab-hyres}
  \end{subfigure}
  \begin{subfigure}[b]{0.24\textwidth}
   \centering
   \includegraphics[width=\textwidth]{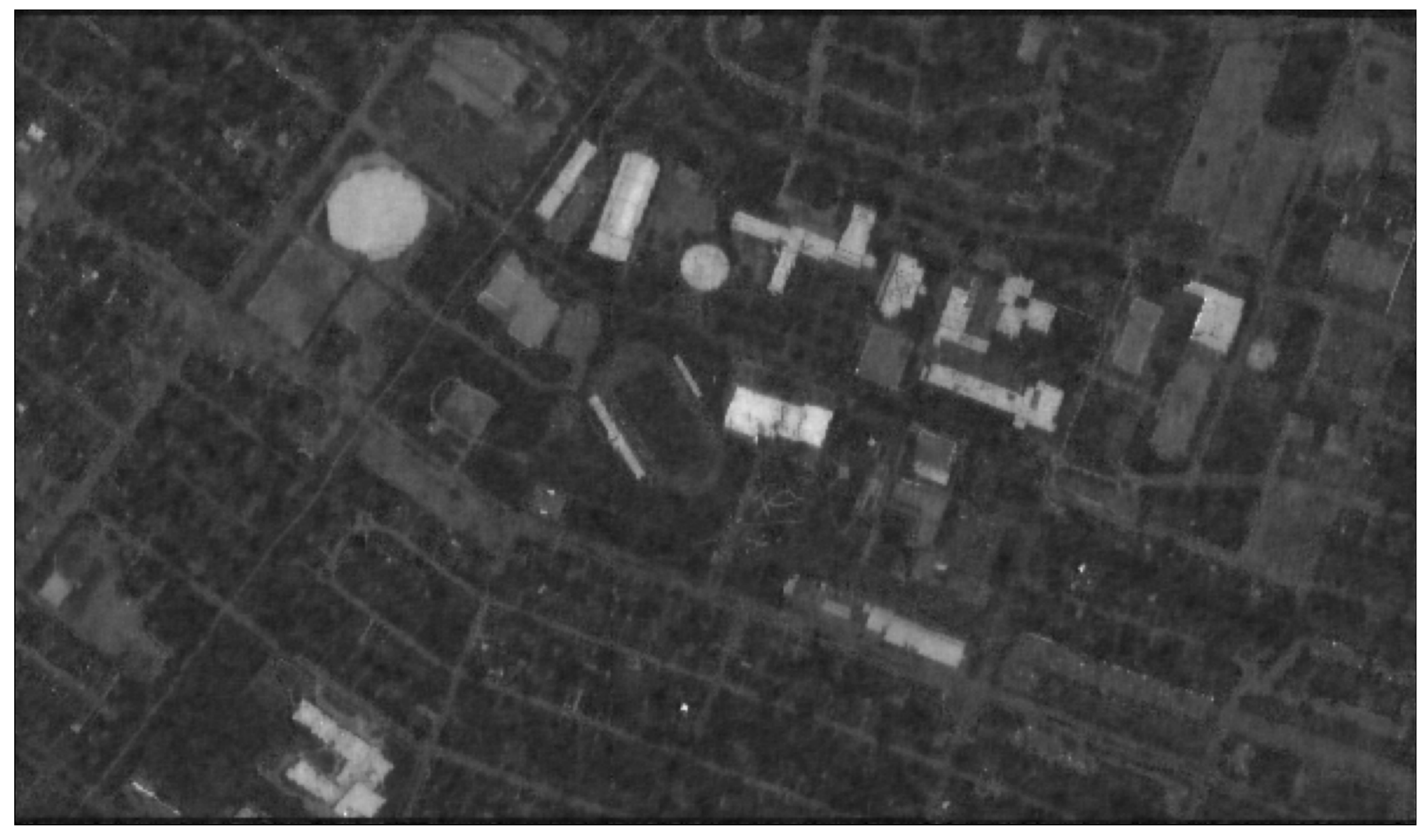}
   \caption{QRNN3D}
   \label{fig:qrnn3d}
  \end{subfigure}
  \caption{Results for Band 11 of the Houston University data cube for HyDe's HyRes method using a GPU (b), the original HyRes implementation (c), and using the HSI inference method described in \Cref{sec:nn-inference} and the QRNN3D archetecture (d). The target HSI (a) had an SNR of \SI{10}{\decibel}.}
  \label{fig:example}
\end{figure*}

HyDe includes a variety of utility functions including normalization methods, noise simulator, and decomposition methods.
These tools are critical to promote interoperability between high-level methods.
The noise-simulating functions include additive Gaussian noise, non-i.i.d. noise, salt-and-pepper noise, deadline noise, and stripped noise.
At a low level, HyDe utilizes GPUs via the PyTorch~\cite{torch} library.

\subsection{High Level Conventional Methods}
\label{sec:trad-methods}

HyDe includes a collection of high-level conventional methods.
Due to space limitations, we encourage the reader to review the each method's detailed explanation in its cited source.
Our implementations attempt to replicate the reference as accurately as possible.
Many of these methods are compatible with PyTorch's automatic differentation method.

We have included First Order Spectral Roughness Penalty Denoising (FORPDN)~\cite{forpdn}, Fast Hyperspectral Denoising (FastHyDe)~\cite{fasthyde}, Fast Hyperspectral Inpainting (FastHyIn) \cite{fasthyde}, Hyperspectral Mixed Noise Removal By $\ell_1$-Norm-Based Subspace Representation (L1HyMixDe)~\cite{l1hymixde}, Automatic Hyperspectral Restoration (HyRes)~\cite{hyres}, Hyperspectral Mixed Gaussian and Sparse Noise Reduction~(HyMiNoR) \cite{hyminor}, Wavelet-Based Sparse Reduced-Rank Regression (WSRRR) \cite{wsrrr}, and Orthogonal Total Variation Component Analysis (OTVCA)~\cite{otvca} in the first release of HyDe.

FORPDN is a full-rank, wavelet-based method which is primarily adapted for removing additive Gaussian noise.
HyRes and WSRRR are low-rank, wavelet-based methods for removing Gaussian noise.
HyRes is parameter-free and WSRRR can be used for extracting features.

HyMiNoR is a low-rank, wavelet-based method build on top of HyRes.
After the Gaussian noise is removed by HyRes, the remaining sparse noise is removed by solving an $\ell_1$-$\ell_1$ optimization problem with a modified split-Bregman technique.

FastHyDe is a low-rank method based on sparse representations.
It uses the HySime~\cite{HySime} and BM3D~\cite{bm3d} methods.
HySime is implemented in HyDe and can utilize GPUs.
However, as BM3D is a closed-source package dependent on NumPy~\cite{numpy}; it can only utilize CPUs.
FastHyDe is primarily for removing Gaussian or Poisson noise.

The only difference between FasyHyDe and FastHyIn is the addition of an inpainting mask to the noisy image before the noise is estimated. 
This allows FastHyIn to remove Gaussian, Poisson, and sparse noise, while filling in missing data.
The L1HyMixDe method is built on top of FastHyIn.
Its stated objective is to exploit a compact and sparse representation of the input's rank and self-similarity characteristics.

OTVCA is a low-rank, feature extraction method which can be used for removing additive Gaussian noise and extracting features.
This method uses a SciPy~\cite{scipy} implementation of split-Bregman optimization to perform the total-variation denoising. Due to dependencies on NumPy and SciPy, OTVCA, FastHyDe, FastHyIn, and L1HyMixDe do not effectively utilize GPUs.

\subsection{Neural Methods}
\label{sec:nn}

We have included three DNN models for denoising: QRNN \cite{qrnn}, MemNet~\cite{memnet}, and HSID-CNN~\cite{hsid-cnn, qrnn}. 
The base implementation of QRNN has 2D and 3D versions, where 2D and 3D refer to the type of convolution used.
MemNet and HSID-CNN cannot use 3D convolutions with the default structure, as they will exhaust the available GPU memory during training.
Furthermore, we implemented a version of MemNet with a trainable HyRes step before the standard network layers, named MenNetRes. 
Pre-trained versions of these methods are available within HyDe.
While these methods do not encompass all of the state-of-the-art methods, they show how DNN methods can be implemented in HyDe. 

\subsubsection{HSI Training Method}
\label{sec:nn-training}

It is common for HSI DNN methods to train on a subset of an HSI and validate on another subset~\cite{ImRes2021}. 
While this may show effective results, it will likely result in the overestimation of the method's general performance.  
Therefore, we have developed a DNN training method which performs well on datasets which are not spatially related to the training dataset.

HyDe's training method uses the ICVL dataset~\cite{icvldataset}. 
This dataset has 201 HSIs with a spatial resolution of $1,392\times 1,300$ and 519 spectral bands.
The images used are downsampled versions containing 31 spectral bands.
The HSIs capture a variety of scenes from a ground level perspective.
This varies from most HSIs which are taken from satellites.
There are 175 images in the training set and 26 in the validation set.

Before an HSI passes through the network, it undergoes multiple transformations.
First, it is randomly scaled and cropped; only the height and width are modified, the bands remain the same.
Second, a number of consecutive bands are chosen at random.
We have observed that the bands selected must be consecutive in the case that the bands are ordered by wavelength.
Third, a random geometric transform is applied to improve the model's generalization.
Fourth, Gaussian noise is added to the HSI until the signal-to-noise ratio (SNR) is at a random level between \SI{20}{\decibel} and \SI{30}{\decibel}.
We found that training at higher SNRs was not beneficial to network performance.
To train a network to remove another type of noise, this step would be swapped out with the desired type of noise generation.
Finally, each band of the HSI is normalized between zero and one before being passed into the DNN.
Normalizing after the noise is applied is critical.
In practice, we do not know the normalization constants for the pure signal.
Therefore, we must normalize the input during training as if it has an unknown level of noise.

\begin{table}[t!]
\caption{Denoising results for HyDe on GPUs and their reference implementations on CPUs. The SNR values are those of the noisy HSI. PSNR is the peak SNR of the denoised HSI. SAM results are in radians. Imp. stands for implementation.\vspace{0.1cm}}
\label{tab:results}
\small
\centering
\begin{tabular}{@{}llc|rrr@{}}
\toprule
Method    & Imp.  & SNR & PSNR & SAM & Time [\si{\second}] \\ \midrule
HSID-CNN     & HyDe  & 20 & 32.789 & 0.280 & 1.214    \\
          &       & 40 & 32.698 & 0.229 & 1.189    \\ \midrule
FastHyDe  & HyDe  & 20 & 50.703 & 0.056 & 91.564   \\
          &       & 40 & 56.643 & 0.041 & 152.367  \\
          & Ref.  & 20 & 53.231 & 0.050 & 231.424  \\
          &       & 40 & 57.622 & 0.039 & 231.661  \\ \midrule
FORPDN    & HyDe  & 20 & 46.788 & 0.091 & 0.539    \\
          &       & 40 & 60.415 & 0.042 & 0.497    \\
          & Ref.  & 20 & 48.212 & 0.077 & 21.799   \\
          &       & 40 & 60.049 & 0.041 & 21.191   \\ \midrule
HyMiNoR   & HyDe  & 20 & 43.142 & 0.097 & 1.169    \\
          &       & 40 & 43.670 & 0.088 & 1.141    \\
          & Ref.  & 20 & 44.864 & 0.091 & 103.575  \\
          &       & 40 & 45.556 & 0.086 & 106.461  \\ \midrule
HyRes     & HyDe  & 20 & 50.577 & 0.065 & 0.317    \\
          &       & 40 & 56.952 & 0.040 & 0.244    \\
          & Ref.  & 20 & 52.527 & 0.053 & 5.734    \\
          &       & 40 & 62.863 & 0.038 & 8.439    \\ \midrule
L1HyMixDe & HyDe  & 20 & 50.779 & 0.056 & 1213.060 \\
          &       & 40 & 54.753 & 0.044 & 616.106  \\
          & Ref.  & 20 & 46.169 & 0.054 & 1847.193 \\
          &       & 40 & 43.281 & 0.044 & 1848.607 \\ \midrule
MemNet    & HyDe  & 20 & 39.755 & 0.175 & 4.533    \\
          &       & 40 & 52.309 & 0.051 & 4.522    \\ \midrule
MemNetRes & HyDe  & 20 & 43.156 & 0.125 & 9.487    \\
          &       & 40 & 56.142 & 0.044 & 9.423    \\ \midrule
OTVCA     & HyDe  & 20 & 50.666 & 0.060 & 49.062   \\
          &       & 40 & 53.995 & 0.044 & 49.120   \\
          & Ref.  & 20 & 48.876 & 0.075 & 38.059   \\
          &       & 40 & 56.697 & 0.039 & 38.112   \\ \midrule
QRNN2D    & HyDe  & 20 & 44.785 & 0.097 & 3.973    \\
          &       & 40 & 44.802 & 0.075 & 3.963    \\ \midrule
QRNN3D    & HyDe  & 20 & 45.651 & 0.085 & 4.604    \\
          &       & 40 & 47.651 & 0.063 & 4.594    \\ \midrule
WSRRR     & HyDe  & 20 & 50.194 & 0.056 & 1.402    \\
          &       & 40 & 55.245 & 0.044 & 1.390    \\
          & Ref.  & 20 & 52.776 & 0.052 & 46.056   \\
          &       & 40 & 56.058 & 0.042 & 46.648   \\ \bottomrule
\end{tabular}
\vspace{-0.5cm}
\end{table}

\begin{table*}[t]
\caption{Energy, PSNR, and GPU memory measurements of HSI denosing methods using 2013 University of Houston dataset. 
The average PSNR change is between HyDe's CPU and GPU implementation.
}
\label{tab:energy}
\small
\centering
\begin{tabular}{@{}l|rrr|rrrr@{}}
\toprule
Method &
  \begin{tabular}[c]{@{}c@{}}HyDe\\ CPU - \si{\kilo\joule}\end{tabular} &
  \begin{tabular}[c]{@{}c@{}}Reference\\ CPU - \si{\kilo\joule}\end{tabular} &
  \begin{tabular}[c]{@{}c@{}}Ratio \\ Ref. / HyDe \end{tabular} &
  \begin{tabular}[c]{@{}c@{}}HyDe\\ GPU - \si{\kilo\joule}\end{tabular} &
  \begin{tabular}[c]{@{}c@{}}Ratio (HyDe) \\ CPU / GPU \end{tabular} &
  \begin{tabular}[c]{@{}c@{}}Avg. PSNR Change\\ CPU v GPU [\si{\percent}]\end{tabular} &
  \begin{tabular}[c]{@{}c@{}}Peak Memory\\ GB \end{tabular} \\ \midrule
HSID-CNN        & 1,602  & --     & --     & 92     & 17.407 & 0.110  & 2.481 \\
FastHyDe     & 3,535  & 7,628  & 2,158  & 3,560  & 0.993  & -0.002 & 3.078 \\
FORPDN       & 316    & 723    & 2.289  & 40     & 7.963  & -0.003 & 6.718 \\
HyMiNoR      & 929    & 3,403  & 3.665  & 105    & 8.851  & -3.590 & 6.521 \\
HyRes        & 146    & 354    & 2.427  & 31     & 4.697  & 0.032  & 5.516 \\
L1HyMixDe    & 26,298 & 54,932 & 2.089  & 26,237 & 1.002  & -0.003 & 6.686 \\
MemNet       & 2,811  & --     & --     & 191    & 14.738 & 0.006  & 4.177 \\
MemNetRes    & 4,599  & --     & --     & 381    & 12.072 & -0.012 & 2.756 \\
OTVCA        & 1,440  & 1,237  & 0.859  & 1,463  & 0.984  & 0.000  & 2.303 \\
QRNN2D       & 6,338  & --     & --     & 155    & 40.823 & -0.014 & 2.284 \\
QRNN3D       & 15,374 & --     & --     & 189    & 81.519 & 0.014  & 2.312 \\
WSRRR        & 449    & 1,510  & 3.365  & 61     & 7.380  & 0.304  & 6.413 \\ \bottomrule
\end{tabular}
\end{table*}

\subsubsection{DNN Inference on Large HSIs}
\label{sec:nn-inference}

As the intermediate states of many DNNs can be exponentially larger than the input to the network, the memory requirements for DNNs can become problematic if the input is large.
This is frequently the case for HSIs. 
Therefore, we use a sliding window approach to denoise large HSIs in sections.
The sections are created by slicing the target HSI both spatially and spectrally.
This method provides accurate results, although it increases the time required by a factor proportional to the number of tiles to be denoised.

\section{Experiments}
\label{sec:exps}

The methods in HyDe are compared with the reference implementations (where available) in three experiments.
The reference implementations for all non-DNN methods (\cref{sec:trad-methods}) are in MATLAB.
The experiments were performed using the Houston 2013 data cube~\cite{houston} at SNRs of \SI{20}{\decibel}, \SI{30}{\decibel}, and \SI{40}{\decibel}.
The SNR of the target HSI was reduced using additive Gaussian noise.
Each experiment had 15 different measurements.
Olympic scoring is used for averaging, i.e., the fastest and slowest runs are disregarded.
For each method we report the computation time, the peak SNR (PSNR), the average spectral angle mapping (SAM) in radians, the peak GPU memory consumption (where applicable), and the total energy consumed during all three experiments.


The experiments were conducted on a single node of the HoreKa supercomputer at KIT and used a single GPU.
The node had four NVIDIA A100 GPUs~\cite{a100}, 72 Intel Xeon Platinum 8368 CPUs, and 512 GB of RAM. 

HyDe's default numerical precision is 32-bit floating point, while the precision of the original implementations is 64-bit floating point.
Reduced precision can negatively effect the accuracy of operations on extremely small scales; a common occurrence for some of these methods.
Furthermore, the low-level IEEE-754 floating-point operations used on GPUs are slightly different than those used on the CPUs.
This can also effect the accuracy of some calculations.
If an HSI resides on the GPU, HyDe's functions will attempt to primarily use that device.
If a specific function cannot use the GPU, it will use the CPU.

\Cref{tab:results} shows the results of the \SI{20}{\decibel} and \SI{40}{\decibel} SNR-in experiments. 
The CPU results for HyDe and the \SI{30}{\decibel} measurements are reserved for a full length manuscript.
For the conventional methods, there is a slight performance difference between HyDe and the original implementations.
This is likely due to a combination of the reduced precision of HyDe's methods and the differences in low-level operations on GPUs and CPUs. 
The best performing conventional methods were HyRes and FORPDN, however these methods generally outperformed the DNN methods.
The best performing neural approaches were QRNN3D and MemNetRes.


An energy-focused comparison of HyDe's methods is shown in \Cref{tab:energy}.
The reference implementations required more energy for all methods.
For many of the methods, HyDe is roughly a factor of two more efficient when only using the CPUs.
This is due to the reduced precision that HyDe uses.
Anything beyond a factor of two is due to algorithmic design choices in HyDe or PyTorch.
The methods which effectively use GPUs are between 4.7x and 81.5x more energy efficient than HyDe's CPU counterparts. 
This increase is on top of the ratios shown in \Cref{tab:energy}'s fourth column.

The PSNR results change when switching to the GPU but this change is typically less than \SI{0.2}{\percent}.
HyMiNoR is the only method which is noticeably effected by the shift as it uses extremely small numbers.

As the amount of GPU memory is often a limiting factor, \Cref{tab:energy} includes the peak GPU memory usage of each method.
There are no methods which utilize more than 6.72 GB of GPU memory.
This is due to the DNN inference approach shown in \Cref{sec:nn-inference}.
Without this approach, all shown DNN methods attempted to allocate more than 40 GB of GPU memory to denoise the target HSI.

\section{Conclusion}

HyDe is a new package with multiple, GPU accelerated, conventional HSI denoising methods.
These methods perform similarly to their reference implementations while consuming nearly 10 times less energy than their references with minor accuracy loss.  

HyDe also provides three basic DNN architectures and a training procedure which produces networks capable of denoising general HSIs.
As HSIs are already large, the intermediate network states of a denoising network can become prohibitively large.
Therefore, we provide a sliding window inference method which denoises the HSI in sections.
These networks do not outperform conventional methods, but they are competitive.
Due to HyDe's PyTorch backend, some of the conventional methods have trainable parameters.

As computing solutions become more complex and are used increasingly, energy efficient solutions must be preferred. 
HyDe's shared utility functions, energy efficient high-level methods, and easy-to-use Python-based interface makes developing new energy efficient HSI methods significantly easier.

\section{Acknowledgements}
This work was performed on the HoreKa supercomputer funded by the Ministry of Science, Research and the Arts Baden-Württemberg and by the Federal Ministry of Education and Research.
This work is supported by the Helmholtz Association Initiative and Networking Fund under the Helmholtz AI platform grant and the HAICORE@KIT partition.
The authors would like to thank the Hyperspectral Image Analysis group and the NSF Funded Center for Airborne Laser Mapping (NCALM) at the University of Houston for providing the data sets used in this study, and the IEEE GRSS Data Fusion Technical Committee for organizing the 2013 Data Fusion Contest.
\bibliographystyle{IEEEbib}
\bibliography{main}

\end{document}